# A Facial Expression Classification System Integrating Canny, Principal Component Analysis and Artificial Neural Network

Le Hoang Thai, Nguyen Do Thai Nguyen and Tran Son Hai, *Member, IACSIT*

*Abstract*— Facial Expression Classification is an interesting research problem in recent years. There are a lot of methods to solve this problem. In this research, we propose a novel approach using Canny, Principal Component Analysis (PCA) and Artificial Neural Network. Firstly, in preprocessing phase, we use Canny for local region detection of facial images. Then each of local region's features will be presented based on Principal Component Analysis (PCA). Finally, using Artificial Neural Network (ANN) applies for Facial Expression Classification. We apply our proposal method (Canny_PCA_ANN) for recognition of six basic facial expressions on JAFFE database consisting 213 images posed by 10 Japanese female models. The experimental result shows the feasibility of our proposal method.

## I. INTRODUCTION

Facial Expression Classification is an interesting classification problem. There are a lot of approaches to solve this problem such as: using K-NN, K-Mean, Support Vector Machine (SVM) and Artificial Neural Network (ANN). The k-nearest neighbor (k-NN) or K-Mean decision rule is a common tool in image classification but its sequential implementation is slowly and requires the high calculating costs because of the large representation space of images.

SVM applies for pattern classification even with large representation space. In this approach, we need to define the hyper-plane for pattern classification [13]. For example, if we need to classify the pattern into L classes, SVM methods will need to specify 1+ 2+ … + (L-1) = L (L-1) / 2 hyper-plane. However, SVM may be errors in the case of the image are not in any classes, because SVM will classify it into the nearest classes based on the calculation parameters.

Another popular approach is using Artificial Neural Network for the pattern classification. Artificial Neural Network will be trained with the patterns to find the weight collection for the classification process [1]. This approach overcomes the disadvantage of SVM of using suitable threshold in the classification for outside pattern. If the patterns do not belong any in L given classes, the Artificial Neural Network identify and report results to the outside given classes.

In this paper, we propose a solution for Facial Expression Classification using Principal Component Analysis (PCA) and Artificial Neural Network (ANN) like below:

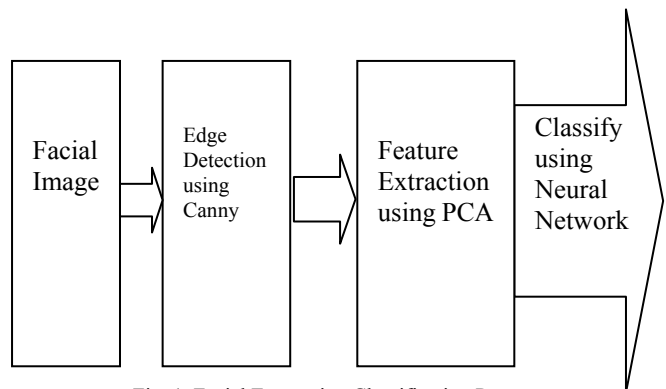

Fig. 1. Facial Expression Classification Process

The facial expression usually expressed in eyes, mouth, brow… Local feature analysis in facial expression is very important for facial expression analysis. Thus, in this approach, we use PCA for local feature extraction and do not apply PCA for whole face. First, we use Canny for local region detection. After that we use PCA to feature extraction in small presenting space.

## II. FACIAL FEATURE EXTRACTION

### A. Canny for local region detection

There are many algorithms for edge detection to detect local feature such as: gradient, Laplacian algorithm and canny algorithm. The gradient method detects the edges by looking for the maximum and minimum in the first derivative of the image. The Laplacian method searches for zero crossings in the second derivative of the image to find edges. The canny algorithm [9,12] uses maximum and minimum threshold to detect edges. The algorithm include following steps:

- Smoothing: using a Gaussian filter to smooth the image to remove noise. A Gaussian filter with σ= 1.4 is shown below:

$$B = \frac{1}{159} \begin{bmatrix} 2 & 4 & 5 & 4 & 2 \\ 4 & 9 & 12 & 9 & 4 \\ 5 & 12 & 15 & 12 & 5 \\ 4 & 9 & 12 & 9 & 4 \\ 2 & 4 & 5 & 4 & 2 \end{bmatrix} \quad (1)$$

- Identifying gradients: First, approximating the gradient in the x- and y-directions by applying the Sobel-operator shown below:

$$H_x = \begin{bmatrix} -1 & 0 & 1 \\ -2 & 0 & 2 \\ -1 & 0 & 1 \end{bmatrix}$$

$$H_y = \begin{bmatrix} 1 & 2 & 1 \\ 0 & 0 & 0 \\ -1 & -2 & -1 \end{bmatrix} \quad (2)$$

Then, applying the law of Pythagoras computes the edge strengths:

$$|G| = \sqrt{G_x^2 + G_y^2}$$
$$|G| = |G_x| + |G_y| \quad (3)$$

Where $G_x$ is the gradient in the x-direction and $G_y$ is the gradient in the y-direction.

The direction of the edges:

$$\theta = \arctan\left(\frac{|G_y|}{|G_x|}\right) \quad (4)$$

- Edge tracking: All local maxima in the gradient image marked as edges, then using double threshold to determine strong edges and weak edges. Remove all edges that are not connected to a strong edge.

In this research, we used Canny algorithm [9,12] to detect local regions for the facial expression features – left and right eyebrows, left and right eyes, and mouth. First, we crop the original image (256x256) into cropped image (85x85) only contain face. After applying histogram equalization, we use canny algorithm for local region detection. Figure 2 shows Facial Feature Extraction Process. Figure 3 shows a sample image, and figure 4 shows the local region detection for the facial features. Figure 5 shows results detected by edge detection using canny algorithm.

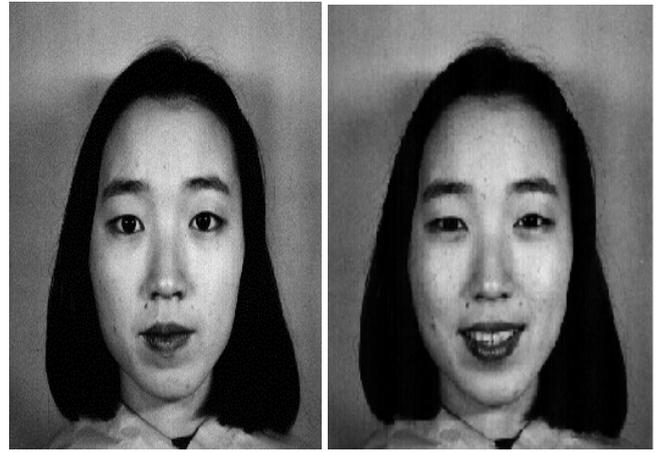

Fig. 3. An Facial Image in JAFEE

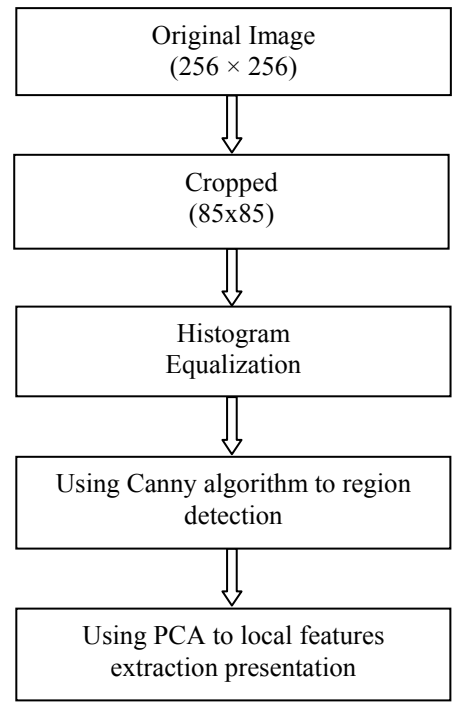

Fig. 2. Facial Feature Extraction Process

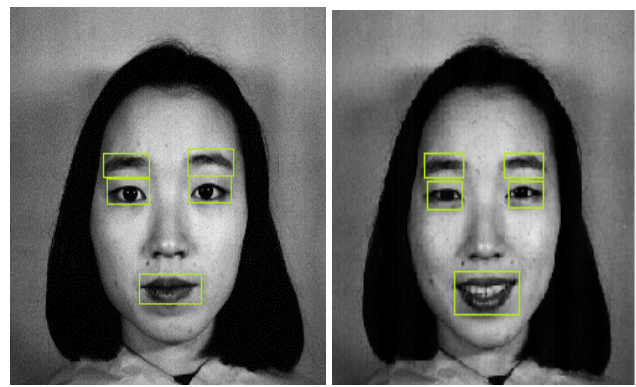

Fig. 4. The local region detection for the facial features

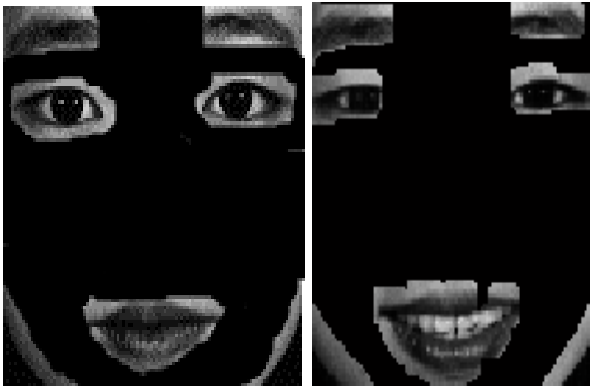

Fig. 5. Results detected by edge detection using canny algorithm

### B. Principal Component Analysis for Facial Feature Extraction

After detected local feature, we used PCA to extract features for left and right eyebrows, left and right eyes, and mouth. These are the vector $v_1$, $v_2$, $v_3$, $v_4$ and $v_5$. Eigenvector is combination of five vectors:

$$V = \{v_1, v_2, v_3, v_4, v_5\} \quad (5)$$

PCA is a procedure that reduces the dimensionality of the data while retaining as much as possible of the variation present in the original dataset. PCA use a linear transformation that converts data from a high dimensional space (x) to a lower dimensional space (y):

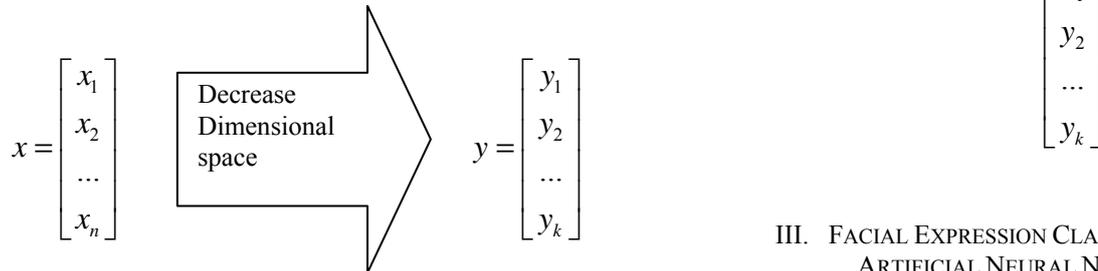

Fig. 6. Decreasing representation space.

$$y = Tx$$
$$y_1 = t_{11}x_1 + t_{12}x_2 + \ldots + t_{1n}x_n$$
$$y_2 = t_{21}x_1 + t_{22}x_2 + \ldots + t_{2n}x_n$$
$$\ldots$$
$$y_k = t_{k1}x_1 + t_{k2}x_2 + \ldots + t_{kn}x_n \quad (6)$$

$$T = \begin{bmatrix} t_{11} & t_{12} & \ldots & t_{1n} \\ t_{21} & t_{22} & \ldots & t_{2n} \\ \ldots & \ldots & \ldots & \ldots \\ t_{k1} & t_{k2} & \ldots & t_{kn} \end{bmatrix}$$

Let $X = \{x_1, x_2, \ldots x_m\}$ are set of Nx1 vectors. The method runs in six steps:

- Computing mean of sets x:

$$\overline{x} = \frac{1}{M}\sum_{i=1}^{M} x_i \quad (7)$$

- Subtract the mean:

$$\varphi_i = x_i - \overline{x} \quad (8)$$

- Set the matrix $A = [\varphi_1 \; \varphi_2 \; \ldots \; \varphi_m]$, then compute:

$$C = AA^T = \frac{1}{M}\sum_{i=1}^{M} \varphi_i \varphi_i^T \quad (9)$$

- Computing the eigenvalues of C:

$$\lambda_1 > \lambda_2 > \ldots > \lambda_n$$

- Computing the eigenvectors of C: $u_1, u_2, \ldots, u_n$

$$x - \overline{x} = y_1 u_1 + y_2 u_2 + \ldots + y_n u_n = \sum_{i=1}^{N} y_i u_i \quad (10)$$

- Dimensionality reduction

$$\hat{x} - \overline{x} = \sum_{i=1}^{K} y_i u_i \quad (11)$$

Where K<<N, $u_i$ are K largest eigenvalues. The representation of $\hat{x} - \overline{x}$ into the basis $u_1, u_2, \ldots, u_k$ is

$$\begin{bmatrix} y_1 \\ y_2 \\ \ldots \\ y_k \end{bmatrix}$$

## III. FACIAL EXPRESSION CLASSIFICATION USING ARTIFICIAL NEURAL NETWORK

In this paper, we use Multi Layer Perceptron (MLP) Neural Network with back propagation learning algorithm.

### A. Multi layer Perceptron (MLP) Neural Network

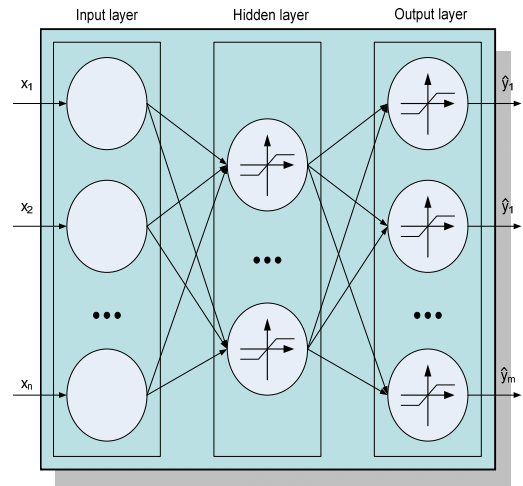

Fig. 7. Multi Layer Perceptron structure

A Multi Layer Perceptron (MLP) is a function

$$\hat{y} = MLP(x, W), \text{ with } x = (x_1, x_2, ..., x_n) \text{ and } \hat{y} = (\hat{y}_1, \hat{y}_2, ..., \hat{y}_m) \quad (10)$$

W is the set of parameters $\{w_{ij}^L, w_{i0}^L\}, \forall i, j, L$

For each unit i of layer L of the MLP. Integration:

$$s = \sum_j y_j^{L-1} w_{ij}^L + w_{i0}^L \quad (12)$$

Transfer: $y_j^L = f(s)$, where

$$f(t) = \frac{1}{1 + e^{-t}} \quad (13)$$

On the input layer (L = 0): $y_j^L = x_j$

On the output layer (L = **L**): $y_j^L = \hat{y}_j$

The MLP uses the algorithm of Gradient Back-Propagation for training to update W.

### B. Structure of MLP Neural Network

MLP Neural Network applies for seven basic facial expression analysis signed MLP_FEA. MLP_FEA has 7 output nodes corresponding to anger, fear, surprise, sad, happy, disgust and neutral. The first output node give the probability assessment belong anger.

MLP_FEA has 200 input nodes corresponding to the total dimension of five feature vectors in V set.

The number of hidden nodes and learning rate $\lambda$ will be identified based on experimental result.

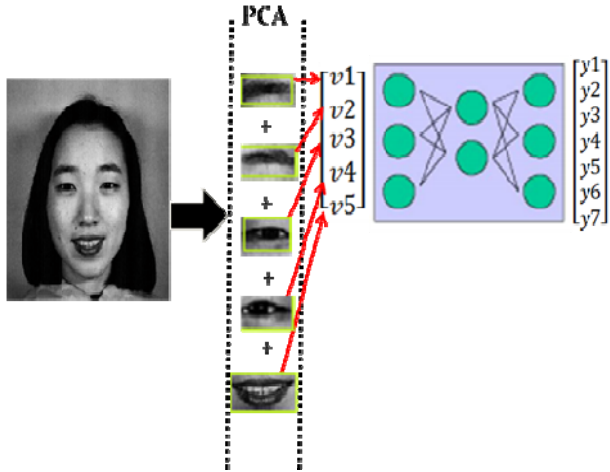

Fig. 8. Structure of MLP Neural Network

TABLE I. OUTPUT NODE CORRESPONDING TO ANGER, FEAR, SURPRISE, SAD, HAPPY, DISGUST AND NEUTRAL

| Feeling | Max |
|---|---|
| Anger | Y1 |
| Fear | Y2 |
| Surprise | Y3 |
| Sad | Y4 |
| Happy | Y5 |
| Disgust | Y6 |
| Neutral | Y7 |

## IV. EXPERIMENTAL RESULT

We apply our proposal method for recognition of six basic facial expressions on JAFEE database consisting 213 images posed by 10 Japanese female models. We conduct the fast training phase (with maximum 200000 epochs of training) with the learning rate $\lambda$ in {0.1, 0.2, 0.3, 0.4, 0.5, 0.6, 0.7, 0.8, 0.9} and the number of hidden nodes in {5, 10, 15, 20, 25} to identify the optimal MLP_FEA configuration. The precision of classification see the table below:

TABLE II. FAST TRAINING WITH 200000 EPOCHS

| Hidden Nodes | Learning rate $\lambda$ | | | | | | | | |
|---|---|---|---|---|---|---|---|---|---|
| | 0.1 | 0.2 | 0.3 | 0.4 | 0.5 | 0.6 | 0.7 | 0.8 | 0.9 |
| 5 | 78.57 | 74.29 | 75.71 | 71.43 | 72.86 | 75.71 | 77.14 | 71.43 | 74.29 |
| 10 | 80.00 | 78.57 | 84.29 | 80.00 | 81.43 | 81.43 | 80.00 | 82.86 | 78.57 |
| 15 | 77.14 | 75.71 | 74.29 | 80.00 | 81.43 | 82.86 | 78.57 | 75.71 | 81.43 |
| 20 | 78.57 | 75.71 | 78.57 | 74.29 | 75.71 | 75.71 | 82.86 | 81.43 | 80.00 |
| 25 | 68.57 | 71.43 | 70.00 | 71.43 | 68.57 | 70.00 | 72.86 | 71.43 | 71.43 |

It is easy to see that the best classification with $\lambda = 0.3$ and the number of hidden nodes = 10. Therefore, we develop ANN with 10 hidden nodes and $\lambda = 0.3$ to apply for recognition of six basic facial expressions.

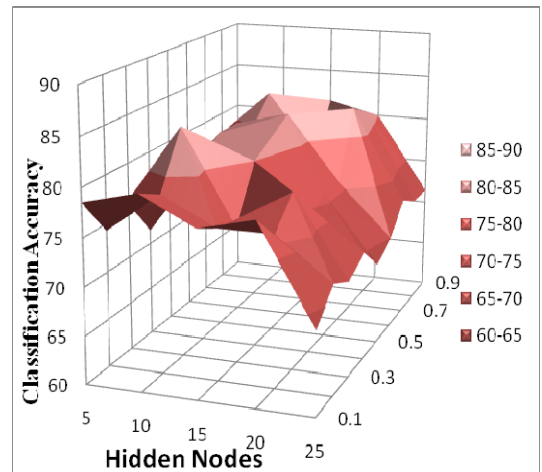

Fig 9. 3D chart of Fast Training with 200000 epochs

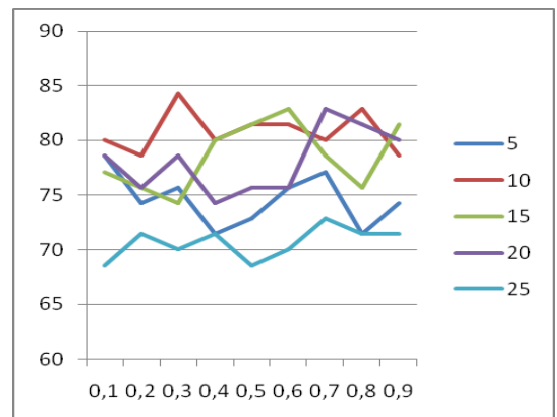

Fig 10. Fast Training with 200000 epochs

Based on the above optimal MLP_FEA configuration, we conduct the training with error = $10^{-7}$ and obtained the result below:

TABLE III. FACIAL EXPRESSION CLASSIFICATION

| Feeling | Correct Classifications | Classification Accuracy % |
|---|---|---|
| Anger | 9/10 | 90 |
| Fear | 8/10 | 80 |
| Surprise | 9/10 | 90 |
| Sadness | 9/10 | 90 |
| Joy | 8/10 | 80 |
| Disgust | 9/10 | 90 |
| Neutral | 8/10 | 80 |

The average facial expression classification of our proposal method (Canny_PCA_ANN) is 85.7%. We compare our proposal methods with Rapid Facial Expression Classification Using Artificial Neural Network [10], Facial Expression Classification Using Multi Artificial Neural Network [11] in the same JAFFE database.

TABLE IV. COMPARATION CLASSIFICATION RATE OF METHODS

| Method | Classification Accuracy % |
|---|---|
| Rapid Facial Expression Classification Using Artificial Neural Networks [10] | 73.3% |
| Facial Expression Classification Using Multi Artificial Neural Network [11] | 83.0% |
| Proposal System (Canny_PCA_ANN) | 85.7% |

This method (Canny_PCA_ANN) improved the Classification Accuracy than Rapid Facial Expression Classification Using Artificial Neural Networks [10] and Facial Expression Classification Using Multi Artificial Neural Network [11] (only used ANN).

Beside, this method does not need face boundary detection process perfect correctly. We used Canny for search local regional (left – right eyebrow, eyes and mouth) directly.

## V. CONCLUSION

In this paper, we suggest a new method using Canny, Principal Component Analysis (PCA) and Artificial Neural Network (ANN) apply for facial expression classification. Canny and PCA apply for local facial feature extraction. A facial image is separated to five local regions (left eye, right eye, left and right eyebrows and mouth). Each of those regions' features is presented by PCA. So that image representation space is reduced.

Instead of using ANN based on the large image representation space, ANN is used to classify Facial Expression based on PCA representation. So the training time of ANN is reduced.

To experience the feasibility of our approach, in this research, we built recognition of six basic facial expressions system on JAFFE database consisting 213 images posed by 10 Japanese female models. The experimental result shows the feasibility of our proposal system.

However, this approach uses ANN for classifying and the number of hidden nodes is identified by experience. It required the high calculating cost for learning process.

REFERENCES

[1] S. Tong, and E. Chang, "Support vector machine active learning for image retrieval", in Proc. ninth ACM international conference on Multimedia, New York, 2001, pp. 107-118.

[2] V. H. Nguyen, "Facial Feature Extraction Based on Wavelet Transform", Lecture Note in Computer Science, Springer: Press, 2009, Vol 5855/2099, pp 30-339.

[3] M. J. Lyons, J. Budynek, S. Akamatsu, "Automatic Classification of Single Facial Images", IEEE Transactions on Pattern Analysis and Machine Intelligence , 1999, Vol 21, pp1357-1362.

[4] Y. Cho and Z. Chi, "Genetic Evolution Processing of Data Structure for Image Classification", IEEE Transaction on Knowledge and Data Engineering, 2005, Vol.17, No 2, pp 216-231.

[5] S. T. Li and A. K. Zan, Hand Book of Face Regconition, Springer: Press, 2005.

[6] C. M. Bishop, Pattern Recognition and Machine Learning, Springer: Press, 2006.

[7] I. Buciu and I. Pista, "Application of non-Negative and Local non Negative Matrix Factorization to Facial Expression Recognition", the 17th International Conference on Pattern Recognition , Patern Recognition, 2004, Vol. 1, pp 288-291.

[8] P. Zhao-yi , Z. Yan-hui , Z. Yu, "Real-time Facial Expression Recognition Based on Adaptive Canny Operator Edge Detection", in Proc. Second International Conference on MultiMedia and Information Technology, 2010, Vol. 1, pp 154-157.

[9] F. Mai, Y. Hung, H. Zhong, and W. Sze, "A hierarchical approach for fast and robust ellipse extraction", IEEE International Conference on Image Processing, 2007, Vol. 5, pp 345-349.

[10] Nathan Cantelmo (2007), "Rapid Facial Expression Classification Using Artificial Neural Networks", Northwestern University [online], Available: http://steadystone.com/research/ml07/ncantelmo_final.doc

[11] L. H. Thai, T. S. Hai, "Facial Expression Classification Based on Multi Artificial Neural Network", in Proc. International conference on Advance Computing and Applications, Vietnam, 2010, pp. 125-133

[12] J. Canny, "A computational approach to edge detection", IEEE Transactions on Pattern Analysis and Machine Intelligence, 1986, Vol.PAMI-8, No 6, pp 679–698.

[13] S. Tong, E. Chang, "Support vector machine active learning for image retrieval", in Proc. 9th ACM international conference on Multimedia , New York, 2001, pp 107-118

**Dr Le Hoang Thai** received B.S degree and M.S degree in Computer Science from Hanoi University of Technology, Vietnam, in 1995 and 1997. He received Ph.D. degree in Computer Science from Ho Chi Minh University of Sciences, Vietnam, in 2004. Since 1999, he has been a lecturer at Faculty of Information Technology, Ho Chi Minh University of Natural Sciences, Vietnam. His research interests include soft computing pattern recognition, image processing, biometric and computer vision. Dr. Le Hoang Thai is co-author over twenty five papers in international journals and international conferences.

**Nguyen Do Thai Nguyen** received B.S degree from University of Pedagogy, Ho Chi Minh city, Vietnam, in 2007. He is currently pursuing the M.S degree in Computer Science Ho Chi Minh University of Science.
From 2008-2010, he has been a lecturer at Faculty of Mathematics and Computer Science in University of Pedagogy, Ho Chi Minh city, Vietnam. His research interests include soft computing pattern recognition, machine learning and computer vision. Mr. Nguyen Do Thai Nguyen is co-author of two papers in the international conferences.

**Tran Son Hai** is a member of IACSIT and received B.S degree and M.S degree in Ho Chi Minh University of Natural Sciences, Vietnam in 2003 and 2007. From 2007-2010, he has been a lecturer at Faculty of Mathematics and Computer Science in University of Pedagogy, Ho Chi Minh city, Vietnam. Since 2010, he has been the dean of Information System department of Informatics Technology Faculty and a member of Science committee of Informatics Technology Faculty. His research interests include soft computing pattern recognition, and computer vision. Mr. Tran Son Hai is co-author of four papers in the international conferences and national conferences.